\documentclass[preprint,5p,times,twocolumn]{elsarticle}
\usepackage{gensymb}
\usepackage{tikz}
\usepackage{graphicx}
\usepackage{multirow}
\usepackage{caption}

\usepackage{lineno}

\usepackage{array}
\usepackage{amsmath,amssymb,stmaryrd}  
\usepackage{booktabs}
\usepackage{multirow}
\usepackage{adjustbox}
\usepackage{float}
\usepackage{algorithm}
\usepackage{algpseudocode}

\journal{Robotics and Autonomous Systems}

\begin{document}

\begin{frontmatter}



\title{Transformation \& Translation Occupancy Grid Mapping: 2-Dimensional Deep Learning Refined SLAM}

\author{Leon Davies}

\affiliation{organization={Department of Computer Science, Loughborough University},
            addressline={Epinal Way}, 
            city={Loughborough},
            postcode={LE11 3TU}, 
            state={Leicestershire},
            country={United Kingdom}}

\author{Baihua Li}


\author{Mohamad Saada}


\author{Simon Sølvsten}

\affiliation{organization={European Center for Risk \& Resilience Studies, University of Southern Denmark},
            addressline={Degnevej 14}, 
            city={Esbjerg},
            postcode={6705}, 
            state={Esbjerg},
            country={Denmark}}

\author{Qinggang Meng}


\begin{abstract}
SLAM (Simultaneous Localisation and Mapping) is a crucial component for robotic systems, providing a map of an environment, the current location and previous trajectory of a robot. While 3D LiDAR SLAM has received notable improvements in recent years, 2D SLAM lags behind. Gradual drifts in odometry and pose estimation inaccuracies hinder modern 2D LiDAR-odometry algorithms in large complex environments. Dynamic robotic motion coupled with inherent estimation based SLAM processes introduce noise and errors, degrading map quality. Occupancy Grid Mapping (OGM) produces results that are often noisy and unclear. This is due to the fact that evidence based mapping represents maps according to uncertain observations. This is why OGMs are so popular in exploration or navigation tasks. However, this also limits OGM's effectiveness for specific mapping based tasks such as floor plan creation in complex scenes. To address this, we propose our novel Transformation \& Translation Occupancy Grid Mapping (TT-OGM). We adapt and enable accurate and robust pose estimation techniques from 3D SLAM to the world of 2D and mitigate errors to improve map quality using Generative Adversarial Networks (GANs). We introduce a novel data generation method via deep reinforcement learning (DRL) to build datasets large enough for training a GAN for SLAM error correction. We demonstrate our SLAM in real-time on data collected at Loughborough university. We also prove its generalisability on a variety of large complex environments on a collection of large scale well-known 2D occupancy maps. Our novel approach enables the creation of high quality OGMs in complex scenes, far surpassing the capabilities of current SLAM algorithms in terms of quality, accuracy and reliability.

\end{abstract}





\end{frontmatter}


\section{Introduction}

\begin{figure*}[ht]
    \centering
       \includegraphics[width=1\linewidth]{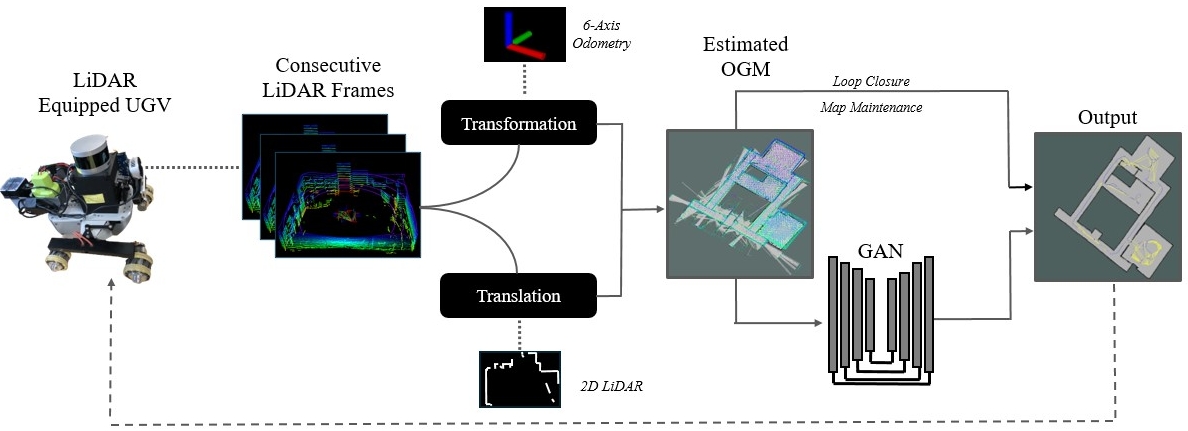}
    \caption{Transformation \& Translation Occupancy Grid Mapping - system overview. Accurate and noise free 2D SLAM from the singular input of consecutive 360\textdegree{} LiDAR scans. Transformation and Translation enables accurate 3D pose estimation algorithms to be relevant for 2D OGM. Error correction and artefact removal is applied through deep learning.}
    \label{fig:mainsteps}
\end{figure*}

Occupancy Grid Mapping (OGM) is an important process in enabling robots and their operators to determine their spatial location and visualise the world around them. In OGM, the world is broken up into grid cells. Each grid cell denotes a state of occupancy corresponding to its position in the map, providing data into nearby occupied, unoccupied and unexplored spaces. This enables robots to navigate autonomously and interact with their environment dynamically, facilitating tasks such as path planning and exploration. 

In recent years LiDAR-odometry algorithms such as LOAM \cite{LOAM}, LeGO LOAM \cite{lego-loam}, and Direct LiDAR-Odometry \cite{dlo} have made significant strides in producing accurate pose estimation and high quality 3D maps, particularly in complex scenes with dynamic movement. This improvement however, has not been paralleled for OGM.

While 2D SLAM systems such as SLAM Toolbox \cite{slam-toolbox} and Hector Mapping \cite{Hector} are well optimised for OGM in small scale steady Unmanned Ground Vehicles (UGVs), they struggle to maintain accuracy in larger environments or when more complex movements are involved. This disparity in accuracy and quality may encourage roboticists to use 3D SLAM over OGM in many situations. This however is not possible for some tasks, one example being floor plan creation. Floor plan creation through OGM enables quick and distance-accurate graphing of buildings. 
OGM can also be advantageous for embedded systems on mobile robots due to the simplicity of 2D OGM compared to handling large 3D point clouds. The complexity of calculations and storage is typically much lower with OGM than with 3D SLAM. This is especially relevant when considering further downstream tasks such as exploration. In which usage of an OGM can be much less computationally expensive than computing directly from point clouds. This is most appropriate in applications where speed and real-time performance are critical or in applications with limited computational resources. 

A limiting factor of OGM is the output quality of the produced map, which is much more susceptible to noise and artefacts compared to 3D representations. This susceptibility arises from the underlying processes of estimation and optimisation that enable SLAM. This may go unnoticed at smaller scales in 3D point clouds, but is much more noticable in OGM. Additional errors and undesirable data may be introduced from factors such as the sensors employed, such as high-dimensional LiDAR scanners, environmental factors like the presence of reflective or transparent materials, or the dynamic motion of robotic movement. Slight errors in pose estimation or gradual drift of odometry can lead to issues such as linear and angular offsets. These adversely affect the accuracy and reliability of the produced maps. These challenges are particularly pronounced when utilising lower-cost, less powerful sensors \cite{lidar-sensors}. However, even with powerful long-range sensors, the presence of new noises, such as accidentally captured observations such as LiDAR capturing data through open doorways or glass, can introduce irrelevant data into the map. This negatively impacts overall map quality. 

Many SLAM algorithms demonstrate success in controlled environments but struggle to produce reasonable results in large, complex indoor areas \cite{slam-errors}. Our specific use-case focuses on autonomous mapping of industrial warehouses and factories that are complex in nature. Improvements and updates to OGM are essential for extending its usability to this area.

Enabling this would effectively allow autonomous systems to handle a broader range of tasks at larger scales. Furthermore, the ability to produce accurate maps without requiring a steady robot allows for more flexible development of innovative robotic systems. 

Mature 2D LiDAR scanners using scan-to-map matching are popular methods for OGM and may yield better results than 3D LiDARs. This, however comes at the cost of a much lower range, and is not optimised for complex movement, which is unfeasible for our use-case.
Accuracy of pose and quality of map can be improved through techniques such as sensor fusion. This can increase the overall cost and power consumption of a robotic system, as opposed to LiDAR odometry. 


Improving the quality and reliability of OGM in large-scale environments with robots navigating complex and dynamic scenes holds significant importance.

The reduction of errors and noise through using 3D LiDAR-odometry for pose estimation in 2D representations has the potential to improve the confidence of robots executing tasks ranging from obstacle avoidance to fully autonomous navigation.

It empowers automated systems to navigate with more confidence and precision, ensuring safer and more efficient operations in challenging and complex conditions. Furthermore, the production of higher quality map outputs produced through OGM paired with deep learning can expand the utility of OGM across various domains. Particularly in sectors where quickly produced maps can prove useful, such as floor plans for facilities management, or the insurance industry. 


Classical SLAM error correction and optimisation techniques include the usage of probabilistic methods \cite{EKF,MCL,gmapping,fastSLAM}, bundle adjustment \cite{bundle}, loop closure \cite{loop-closure} and graph-based approaches \cite{graphSLAM}. These optimise mapping and localisation during the SLAM process and are fundamental to modern day SLAM systems.
Evidence based occupancy mapping techniques \cite{32,33,34} build OGMs based on probabilistic models by integrating LiDAR data while accounting for uncertainties. These techniques are susceptible to noise, artefacts and slight irregularities, especially in large scenes and complex situations. A small inaccuracy in pose estimate can drift over time and create an unreadable map, many 2D SLAM OGM algorithms have this problem. These challenges have become accepted as normal within OGM.

The disparity in accuracy between 2D and 3D SLAM motivates our approach to adapt and enable state-of-the-art 3D LiDAR odometry calculations for 2D OGM, effectively modernising 2D SLAM. We do this through our Transformation and Translation stage in our SLAM. 

Many SLAM errors, such as linear/angular error and noise caused by transparent or reflective materials, can be challenging to detect and resolve using traditional programming methods but are easily identified in samples observed and resolved by humans. This motivates our proposed deep learning-based error correction approach. When considering the usage of deep learning to improve SLAM, most research focuses on either visual SLAM \cite{dl-for-vslam, dl-for-vslam2} or specific fundamentals such as loop closure \cite{loop-closure-for-slam}. There has been little work done on OGM, likely due to the scarcity of data.

We draw inspiration from recent advances in Image-To-Image (I2I) translation GANs \cite{cyclegan, cut, qsa} for tasks such as image resolution upscaling \cite{srgan} and image deblurring and denoising \cite{denoise}. We train models capable of achieving similar mapping functions for completion of partially mapped regions and removal of unwanted data in OGM. This produces mapping results appropriate for wider use in tasks such as floor plan creation. 
Developing a deep learning-based solution requires a large amount of diverse and realistic data to enable generalisation of models to real-world scenarios. Publicly available data for this does not currently exist. To alleviate this, we introduce a new dataset through our novel data generation process, using DRL to mass produce data.

The contribution for our work is summarised as:
\begin{itemize}
    \item Transformation and Translation: Enabling 3D pose estimation algorithms to produce 2D representations of point cloud data, facilitating high quality and reliable 2D OGM in large complex scenes with dynamic movement. 
    \item Novel Deep Learning SLAM Observation Completion and Error Deletion: Capable of detecting and removing sensor noise, irrelevant data and artefacts, while also completing partially mapped regions, partial observations and realigning angular/linear offsets.
    \item Novel DRL Data Generation Process: Leveraging the usage of DRL to mass produce samples of high quality, realistic 2D SLAM errors at scale, thereby enabling future deep learning methodologies for the improvement of 2D OGM. 
\end{itemize}

\section{Related Work}

\subsection{Simultaneous Localisation and Mapping}

Traditional methods of particle filter based SLAM such as Monte Carlo Localisation (MCL) \cite{MCL}, Extended Kalman filters (EKF) \cite{EKF}, and Rao-Blackwellized filtering \cite{gmapping} are widely used to enable real-time pose estimation and map construction in SLAM. MCL represents the posterior distribution of the current pose and map as a set of particles. proving reliable in complex environments with non-linear and non-Gaussian noise distributions. EKF applies non-linear motion and measurement models, making it successful in SLAM problems with Gaussian noise assumptions. FastSLAM \cite{fastSLAM} and Gmapping \cite{gmapping} are particle filter based 2D SLAM algorithms. They have both proven reliable when using multiple sensors on steady UGVs. Gmapping is the algorithm used to build the maps in the Radish dataset \cite{radish}. Hector SLAM \cite{Hector} is a 2D SLAM which calculates pose based on purely LiDAR data, it proves success on less steady robots but has no loop closure and is prone to error matching with complex, rapid turns. Cartographer \cite{Cartographer} is a well known 2D SLAM algorithm and produces reliable high quality OGMs, but requires multiple sensors. SLAM toolbox \cite{slam-toolbox} is another notable 2D SLAM, however it struggles to produce reliable results in large environments with complex robotic movement. 

Graph-based SLAM \cite{graphSLAM} offers an optimisation framework that interprets SLAM as a graph optimisation problem. Nodes represent poses, and edges represent constraints between sensors, motions, and poses. The map and trajectory are estimated through nonlinear least squares optimisation. Karto \cite{karto} is a graph optimisation SLAM which attempts to solve sparse decoupling.

Feature-based methods, such as ORB SLAM \cite{ORB}, extract distinctive features as key points in observations. Mapping and localisation are estimated by tracking the shift of key features across observations. Iterative Closest Points (ICP) \cite{icp-old,gicp} is a commonly used scan matching algorithm for LiDAR data. ICP operates by aligning matching points in point clouds by minimising the distance between them. The LOAM \cite{LOAM} algorithm, and its variants \cite{lego-loam, f-loam} are well known examples of multi sensor loosely coupled SLAM systems. LOAM operates by leveraging geometric information from LiDAR data to assume changes in odometry coupled with six-axis IMU to obtain prior pose data, strengthening accuracy. 

Modern day 3D LiDAR-odometry SLAM algorithms such as F-LOAM \cite{f-loam}, HDL-graph SLAM \cite{hdl} and Direct LiDAR-odometry \cite{dlo} are able to produce accurate pose estimates and high quality 3D maps in large environments with complex movement. However. None of these methods are built to produce 2D representations or detect artefacts at scale, assume completed regions or remove irrelevant data. Our method attempts to achieve this. 

\subsection{Deep Learning \& SLAM}
Deep learning has been applied to loop closure within SLAM systems which in turn aids with scan filtering and point cloud registration, to realign angular and linear offsets produced by drifted odometry. 
Superglue \cite{superglue} is an attention-based system that utilises graph neural networks to find corresponding patches between sets of local features for front-end feature extraction. This is applied to the task of pose estimation for indoor and outdoor SLAM on visual data.  

The method proposed in \cite{dlc} aims to decrease drifts in odometry by utilising end-to-end learning-based models for noise reduction and loop closure in 3D LiDAR SLAM. These methods employ traditional EKF odometry estimation but use neural networks for learning-based loop closure and learning based denoising. Boasting impressive results in error reduction for real-time SLAM in outdoor large scale environments.
LPD-NET \cite{LPD-NET} extracts discriminative and generalisable global descriptors. This is done through local feature extraction from point cloud data. local features are processed through neural networks for graph-based neighborhood aggregation and global descriptors. They argue the usage of their model within loop closure. Overlap-Net \cite{OVERLAP-NET} offers a similar approach for loop closure detection through deep neural networks. Their models produce a predicted overlap region and estimated yaw between two point clouds. 



\section{Methodology}
\begin{figure*}
    \centering
    \includegraphics[width=\linewidth]{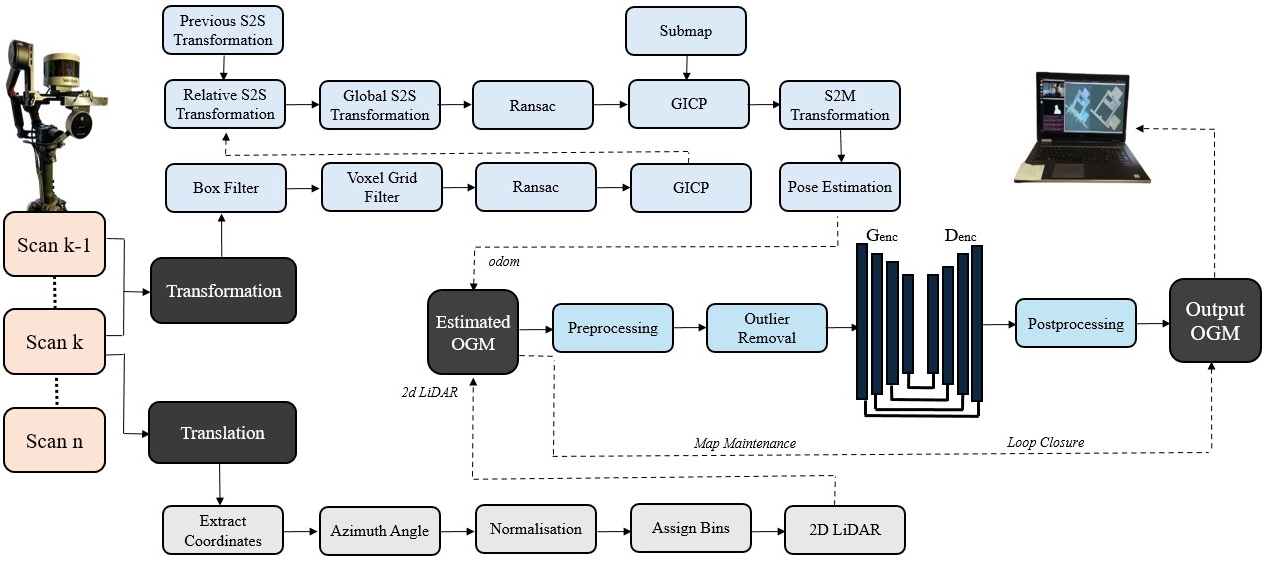}
     \caption{TT-OGM system diagram: Producing a high quality and accurate occupancy grid map in real-time.
     Main stages: 1) Input of consecutive LiDAR point clouds are translated to 2D LiDAR while the transformation between them in calculated through Scan-to-Scan (S2S) transforms and Generalised-iterative closest point (GICP). The position within the world is converted from LiDAR coordinate system using a Scan-To-Map (S2M) transformation. Ransac, box and voxel filters are used for outlier rejection and data decomplexification. 
     2) An estimated OGM is constructed with the pose produced by the transformation branch and the 2D LiDAR produced by the translation branch. 3) The OGM is filtered and processed to prepare it for the GAN model. 4) The OGM is cleaned by the Generator in the GAN model 5) The output is post processed into a clean OGM and provided to the robot or operator.}
     \label{fig:network_diagram}
\end{figure*}
\subsection{Overview of TT-OGM}
Our Translation and Transformation-OGM enables accurate pose estimation using 3D LiDAR to produce OGMs in complex environments characterised by dynamic movement. Additionally, We integrate a novel GAN-based observation completion and error deletion module into our SLAM framework. This module is trained on high quality examples of erroneous occupancy grid maps, built through our DRL data generator. Our module enables real-time detection and resolution of various common SLAM errors, which is beneficial for mapping based tasks. A diagram of our system pipeline is available in Figure. \ref{fig:network_diagram}.

\subsection{Transformation and Translation}
Our SLAM system relies on a sequential input of 360\textdegree{} LiDAR point clouds $P_{(k,k+1, k..)}$, where each $k$ represents a scan index. $P_k$ captures 3D spatial information in the form of a point cloud. These point clouds undergo a two-step simultaneous process: Transformation and Translation. Transformation calculates 6 DOF egomotion, while Translation generates a 2D representation of the point cloud.

\subsubsection{Transformation}

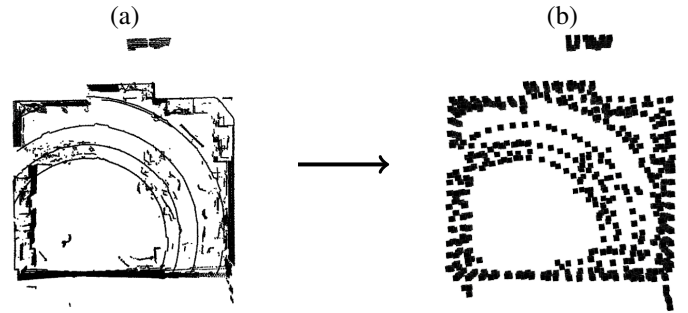
\begin{figure}[H]
    \centering
    \begin{minipage}{0.35\columnwidth} 
        \centering
        \text{(a)}
        \includegraphics[width=\linewidth]{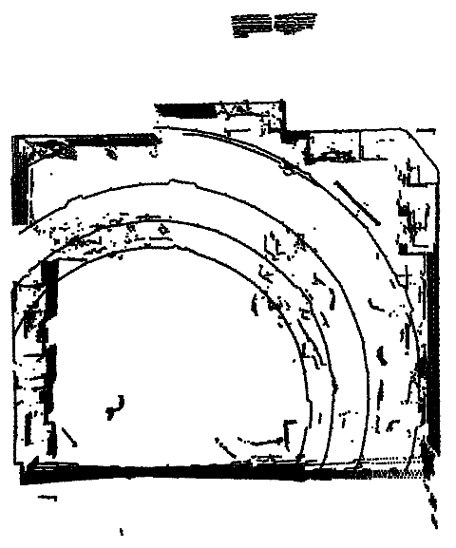}
    \end{minipage}
    \hfill
    \begin{tikzpicture}
        \draw[->, thick, line width=1.5pt, scale=0.8] (0,0) -- (1.5,0); 
    \end{tikzpicture}
    \hfill
    \begin{minipage}{0.35\columnwidth} 
        \centering
        \text{(b)}
        \includegraphics[width=\linewidth]{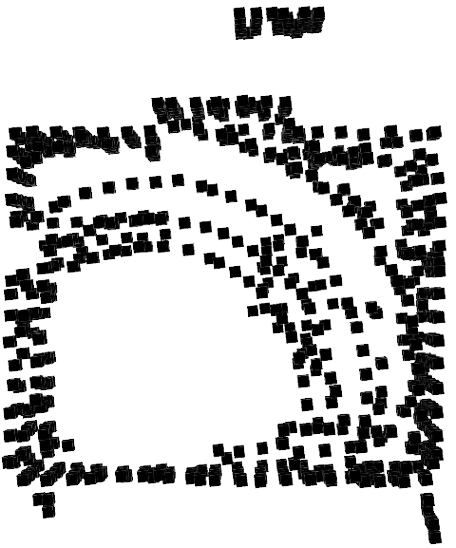}
    \end{minipage}

    \caption{Results of box filter and voxel grid filtering incoming point clouds. (a) Raw point cloud data. (b) Filtered point cloud.}
    \label{fig:pcl_processed}
\end{figure}

We employ a traditional GICP approach \cite{gicp} and adapt \cite{dlo} without IMU or mapping components to calculate the transformation between two, temporally adjacent, point clouds $K$ and $K-1$. Initially, each incoming scan undergoes preprocessing using a box filter to eliminate data points associated with the scanning apparatus or robotic system. We remove all voxels within a $1m^3$ radius from the origin of the point cloud. Subsequently, a 3D voxel grid filter with a resolution of $0.25m$ is applied to downsample the point cloud, reducing it from $30,000$ voxels to $7,500$ voxels. Each voxel corresponds to a volume of $0.25m \times 0.25m \times 0.25m$. A comparison of raw and processed point cloud is available in Figure. \ref{fig:pcl_processed}.


We define the source point cloud ($P^{s}{k}$) as $k$ and the target point cloud ($P^{t}{k}$) as $k-1$. We perform scan-to-scan matching with $P^{s}{k}$ and $P^{t}{k}$ in the LiDAR coordinate system $\mathcal{L}$ to calculate the relative transform $\hat{X}^{\mathcal{L}}_{k}$ using GICP. The residual error $\mathcal{E}$ is computed using Equation \ref{eq:GICP}. In which \( \mathcal{C} \) represents the estimated covariance matrix, and \( d_i=p_i^{\mathrm{t}}-\mathbf{X}_k^{\mathcal{L}} p_i^{\mathrm{s}} \), \( p_i^{\mathrm{s}} \in \mathcal{P}_k^{\mathrm{s}} \), \( p_i^{\mathrm{t}} \in \mathcal{P}_k^{\mathrm{t}} \), for all \( i \). The initial state of \( \tilde{\mathbf{X}}_k^{\mathcal{L}} \) is set to the identity matrix \( \mathbf{I} \).

\begin{align}
   \mathcal{E}\left(\mathbf{X}_k^{\mathcal{L}} \mathcal{P}_k^{\mathrm{s}}, \mathcal{P}_k^{\mathrm{t}}\right)=\sum_i^N d_i^{\top}\left(\mathcal{C}_{k, i}^{\mathrm{t}}+\mathbf{X}_k^{\mathcal{L}} \mathcal{C}_{k, i}^{\mathrm{s}} \mathbf{X}_k^{\mathcal{L}^{\top}}\right)^{-1} d_i
   \label{eq:GICP}
\end{align}

\noindent To produce the relative transform between \( P^{s}{k} \) and \( P^{t}{k} \), we minimise the residual error as shown in Equation \( \ref{eq:rel_trans} \). 

\begin{align}
    \hat{\mathbf{X}}_k^{\mathcal{L}}=\underset{\mathbf{X}_k^{\mathcal{L}}}{\arg \min } \mathcal{E}\left(\mathbf{X}_k^{\mathcal{L}} \mathcal{P}_k^{\mathrm{s}}, \mathcal{P}_k^{\mathrm{t}}\right)
    \label{eq:rel_trans}
\end{align}

\noindent We generate a local submap $S_k$ adapting the method proposed in \cite{dlo} to obtain optimised and consistent motion estimates for determining the transformation between $\mathcal{L}$ and the world coordinate system $\mathcal{W}$. This submap aids in estimating the pose within the world frame $\mathbf{X}_k^{\mathcal{W}}$:

\begin{align}
\hat{\mathbf{X}}_k^{\mathcal{W}}=\underset{\mathbf{X}_k^{\mathcal{W}}}{\arg \min } \mathcal{E}\left(\mathbf{X}_k^{\mathcal{W}} \mathcal{P}_k^{\mathrm{s}}, \mathcal{S}_k\right)
\label{eq:s2m}
\end{align}


\subsubsection{Translation}

We construct a 2D data frame, denoted as $Q_k$, from the 3D point cloud $P_{k}$, ensuring consistency with the pose $\mathbf{X}_k^{\mathcal{W}}$. For each voxel in the point cloud, we extract its Cartesian coordinates $x$ and $y$, along with the intensity $i$. Subsequently, we calculate the Azimuth angle $\theta$ using the formula $\theta = \text{atan2}(y, x)$, which is normalised between $\pi$ and $-\pi$. The distance from each point to the sensor is then computed using the Euclidean distance. 


\begin{figure}[H]
    \centering
    \includegraphics[width=0.75\linewidth]{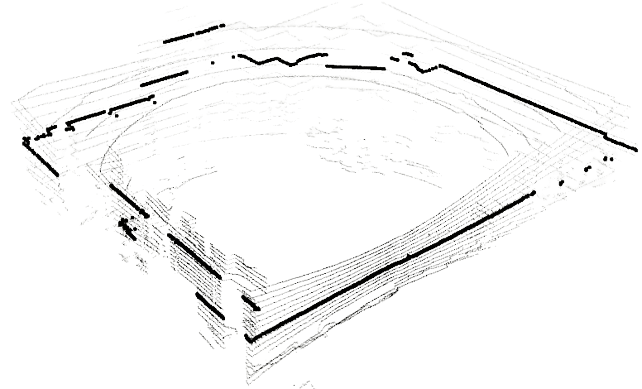}
    \caption{3D LiDAR point cloud and translated 2D representation overlapped.}
    \label{fig:lidar_overlapped}
\end{figure}
    
The 2D data structure is constructed by assigning each point to a specific bin based on its Azimuth angle. 

The resulting $Q_k$ effectively represents a 2D projection of the 3D point cloud $P_{k}$, aligned with the pose $\mathbf{X}k^{\mathcal{W}}$. This alignment enables the creation of a OGM using $Q_k$ wherein the precise odometry calculated from $P{k}$ is accurately reflected in 2D space. A overlay of $P_{k}$ and $Q_{k}$ is shown in Figure. \ref{fig:lidar_overlapped}

Given a 3D point cloud $P$ at time step $k$, with each point $p_i = (x_i, y_i, z_i, i_i) $, where $x_i, y_i, z_i$ are the Cartesian coordinates and $i_i$ is the intensity, the projection from a 3D point cloud onto a 2D plane to construct a 2D data frame $Q_k$ is expressed as pseudocode in Algorithm. \ref{alg:2d-3d}.

\begin{algorithm}[H]
\caption{Convert 3D $P_k$ to 2D $Q_k$}
\label{alg:2d-3d}
\begin{algorithmic}[1]
\State \textbf{Input:} $P_k$
\State \textbf{Output:} $Q_k$
\Function{convert\_lidar}{point\_cloud}
\State num\_points $\gets$ len(point\_cloud)
\State Bins $\gets \sqrt{\text{num\_points}}$ \quad 
\State $Q_k \gets$ [list() for $i$ in range(int(Bins))]
\For{each point in point\_cloud}
\State $(x, y, z, intensity) \gets$ point
\State $\theta \gets \text{atan2}(y, x)$
\If {$\theta > \pi$}
\State $\theta \gets \theta - 2\pi$
\ElsIf {$\theta < -\pi$}
\State $\theta \gets \theta + 2\pi$
\EndIf
\State distance $\gets \sqrt{x^2 + y^2}$
\State bin\_index $\gets \left\lfloor \frac{\theta + \pi}{2\pi} \times \text{int}(Bins) \right\rfloor$
\State $Q_k$[bin\_index].append((distance, intensity))
\EndFor
\State \Return $Q_k$
\EndFunction

\end{algorithmic}
\end{algorithm}
   
\subsection{Occupancy Grid Mapping}

We represent an OGM as a matrix denoted by $M$, where each element $M_{ij}$ of the matrix indicates the occupancy status or likelihood of occupancy at grid cell $(i, j)$. The matrix $M$ is defined as: 
$M = \{ M_{ij} \mid M_{ij} \in [0,255] \text{ for } 1 \leq i \leq H, 1 \leq j \leq W \}$ 

where $M_{ij} = 0 $ indicates that cell $(i, j)$ is unoccupied,
 $M_{ij} = 100 $ indicates that cell $(i, j)$ is occupied and $M_{ij} = 255 $ indicates that cell $(i, j)$ is unexplored. Here, $H$ and $W$ represent the dimensions of Height, and Width.

With newly created 2D LiDAR $Q_k$ and pose estimate $\mathbf{X}_k^{\mathcal{W}}$ 
we build an estimated OGM, $OGM_e$ through evidence based mapping. We construct a pose graph through plotting odometry over time representing trajectory. This pose graph is used for improving localisation adapted from \cite{rtab} along with other SLAM fundamentals including loop closure and map maintenance. These processes are applied at this stage to $OGM_e$ before any changes are made to clean the map with the GAN model. Application of these after modification of the OGM is likely to cause issues which could degrade mapping and localisation quality. $OGM_e$ now represents a highly accurate OGM created through the high dimensional 3D LiDAR. This can be used for further localisation tasks such as navigation. The next section will discuss usage of our GAN to produce a higher quality output for purely mapping based tasks.


\subsection{Occupancy Grid Cleaing}

Estimated $OGM_e$ is taken through a process of filtration to improve its clarity. During this process, occupancy grid cells are marked as outliers and converted back into unexplored cells if they fall into a defined range of occupation intensity. Grid cells with a high or low intensity are adjusted towards the closest value of complete occupancy $M_{ij} = 100$, complete vacancy $M_{ij} = 0$ or completely unexplored $M_{ij} = 255$. We note that this process damages the ground truth occupancy grid values which is why we encourage the usage to be for soley mapping related tasks.

Given the estimated occupancy grid map $OGM_e$ and the intensity values $i_{ij}$ derived from $Q_k$. 
Learnable parameters: $T_1 = 0.12 $ (high intensity threshold), $T_2 = 0.93 $ (lower bound of intensity occupation), $T_3 = 0.96 $ (upper bound of intensity of occupation). Occupation values: $C_1 = 0 $ (unoccupied), $C_2 = 100$ (occupied), $C_3 = 255 $ (unexplored).
For each grid cell $(i, j)$ in $OGM_e$, the filtration process is defined as:

\[
M_{ij} = 
\begin{cases} 
C_1 & \text{if } I_{ij} < T_1 \\
C_2 & \text{if } T_2 \leq I_{ij} \leq T_3 \\
C_3 & \text{if } I_{ij} > T_3 \, \text{or} \, T_1 \leq I_{ij} < T_2 \\
\end{cases}
\]

\vspace{10pt}

\begin{figure}[ht]
    \centering
    \includegraphics[width=1\linewidth]{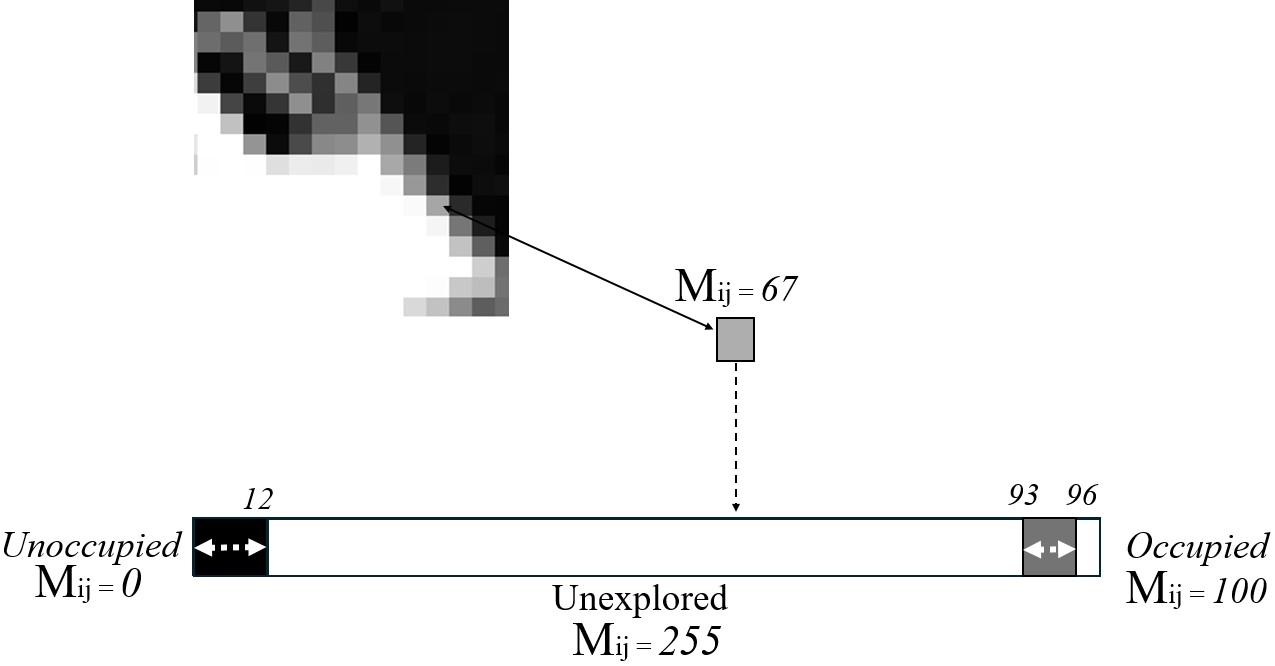}
    \caption{Diagram of filtering operation based on occupancy intensity.}
    \label{fig:filter_diagram2}
\end{figure}

We interpret cells that have a low intensity, $M_{ij} = T_3 , \text{or} , T_1 \leq I_{ij} < T_2$, as having low confidence of occupation or vacancy. These cells have a low probability of occupancy value and can degrade the quality of the OGM. Therefore, cells that fall into this range are marked as outliers and converted back into unexplored cells. Conversely, cells with higher intensities ($M_{ij} < T_1$ or $T_2 \leq I_{ij} \leq T_3$) offer more valuable insights and are accurately represented on the OGM. These are adjusted towards the closest value of complete occupancy or complete vacancy. This improves the readability of the map. Threshold values are treated as learnable parameters, optimised by the performance of our GAN model. Values $M_{ij} > 97$ are treated as unexplored cells despite having a very high intensity of occupation. This is due to a high amount of erroneous data holding occupation values in this range. Removal of these is only possible due to our GAN model. A diagram demonstrating this filtration is available in Figure. \ref{fig:filter_diagram2}.

The filtered $OGM_e$, from which low intensity cells are removed and high intensity cells rounded up, undergoes an additional refinement process of floating point removal.

In this process, pixels that have two or fewer adjacent pixels of the same value are transformed back into unexplored cells. 

\begin{equation}
    Cell(x, y) = 
    \begin{cases}
        Cell(x, y) & \text{if } \sum_{i=1}^{4} \delta(p_i, Cell(x, y)) \geq 2 \\
        255 & \text{if }\sum_{i=1}^{4} \delta(p_i, Cell(x, y)) \leq 2 \\
    \end{cases}
\end{equation}

\noindent In which $p_i$ values represent neighboring cells and $\delta(p_i, \text{Cell}(x, y))$ is a function that returns 1 if a chosen $p_i$ value is the same as the input cell, and 0 otherwise. The value 255 represents unexplored cells.





In these filtering processes, many 'good' pixels with low confidence or lack of neighboring pixels are eliminated, such as some pixels within planar lines collected from slightly reflective materials. It may also seem counterproductive to remove cells within the $M_{ij} > 93$ range; however, this filtering works in tandem with our GAN model. Our GAN is trained to identify and reconstruct the map based on the remaining reliable data. This is enforced through the data used to train the GAN. An abstract way of interpreting this is that the GAN model has insight into what 75,000 different OGMs of buildings look like, both in erroneous with broken segments and reconstructed pixel-perfect variants. The GAN learns the mapping between these, stored in the model's weights.

\begin{figure}[H]
    \centering
    \begin{minipage}{0.4\columnwidth} 
        \centering

        \text{(a)}
        \includegraphics[width=\linewidth]{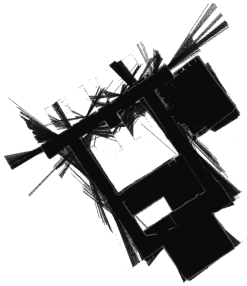}
    \end{minipage}
    \hfill
    \begin{tikzpicture}
        \draw[->, thick, line width=1.5pt, scale=0.8] (0,0) -- (1.5,0); 
        \end{tikzpicture}
    \hfill
    \begin{minipage}{0.4\columnwidth} 
        \centering

        \text{(b)}
        \includegraphics[width=\linewidth]{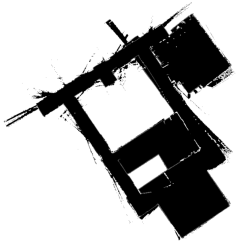}
    \end{minipage}
    
    \vspace{10pt} 
    
    \caption{Results of filtering a occupancy grid map. (a) Unfiltered OGM. (b) Filtered occupancy grid map.}
    \label{fig:filtering_comparison}
\end{figure}
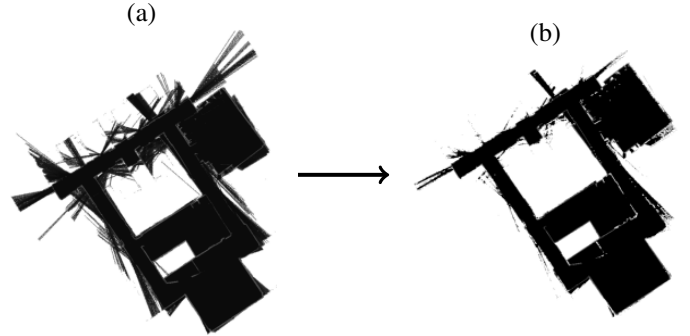

While these filters present no significant challenges for mapping tasks, they would cause significant issues if applied to a traditional 2D OGM. Removing low-intensity cells might inaccurately flag important occupied cells as false positives. These inaccuracies could adversely affect tasks such as path planning or obstacle avoidance. Figure. \ref{fig:filtering_comparison} demonstrates the effect of these filters on a sample of real-world data.



The grid map is now represented with 3 distinct values determining, occupied $M_{ij} = 100$, unoccupied $M_{ij} = 0$ and unexplored cells $M_{ij} = 255$. This process not only improves the clarity of the grid map, but also improves the performance of the GAN model by aligning the values within its training data. %

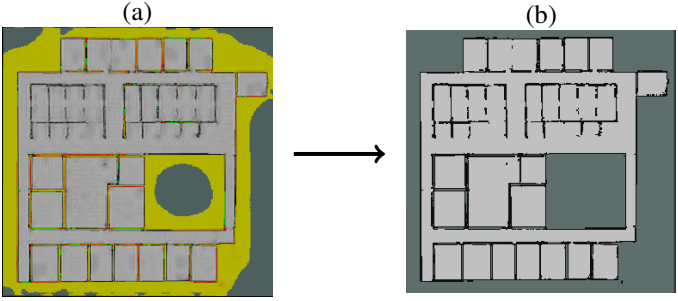
\begin{figure}[H]
    \centering
    \begin{minipage}{0.4\columnwidth} 
    \centering
        \text{(a)}
        \includegraphics[width=\linewidth]{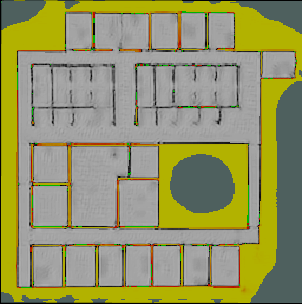}
    \end{minipage}
    \hfill
    \begin{tikzpicture}
        \draw[->, thick, line width=1.5pt, scale=0.8] (0,0) -- (1.5,0); 
    \end{tikzpicture}
    \hfill
    \begin{minipage}{0.4\columnwidth} 
    \centering
        \text{(b)}
        \includegraphics[width=\linewidth]{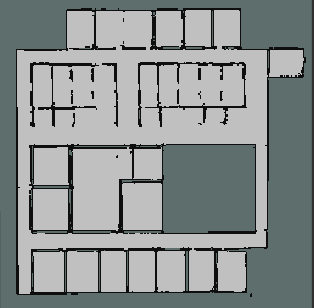}
    \end{minipage}
    
    \vspace{10pt} 
    
    \caption{A demonstration of the necessity of output filtering GAN model predictions. The pixel values of cells in these two examples are nearby but produce very different values within an OGM.(a) An unfiltered OGM produced by the GAN. (b) A filtered OGM. Visualised in ROS}
    \label{fig:filtering_demo}
\end{figure}

At inference, we also apply a output filter to the 'cleaned' prediction as produced by the GAN model. This is necessary as the GAN may make slight changes in cell value. These changes would usually go unnoticed in a RGB image, however the occupancy grid datatype has much more importance and weight on the values of cells, therefore we filter cell values towards their closest value of occupation. A demonstration of the need for this is shown in in Figure. \ref{fig:filtering_demo}.

Given the OGM, $OGM_f$, produced by the GAN model with the intensity values $i_{ij}$, we define the output filtration process similarly to the input filtration process. Learnable parameters: $T_1 = 0.21 $ (high intensity threshold), $T_2 = 0.86 $ (lower bound of intensity occupation). Occupation values: $C_1 = 0 $ (unoccupied), $C_2 = 100$ (occupied), $C_3 = 255 $ (unexplored). For each grid cell $(i, j)$ in $OGM_f$, the output filtration process is defined as:

\[
M_{ij} = 
\begin{cases} 
C_1 & \text{if } I_{ij} < T'_1 \\
C_2 & \text{if } T'_1 \leq I_{ij} \leq T'_2 \\
C_3 & \text{if } I_{ij} > T'_2 \\
\end{cases}
\]

\subsection{GAN Model}

\begin{figure*}
    \centering
    \includegraphics[width=1\linewidth]{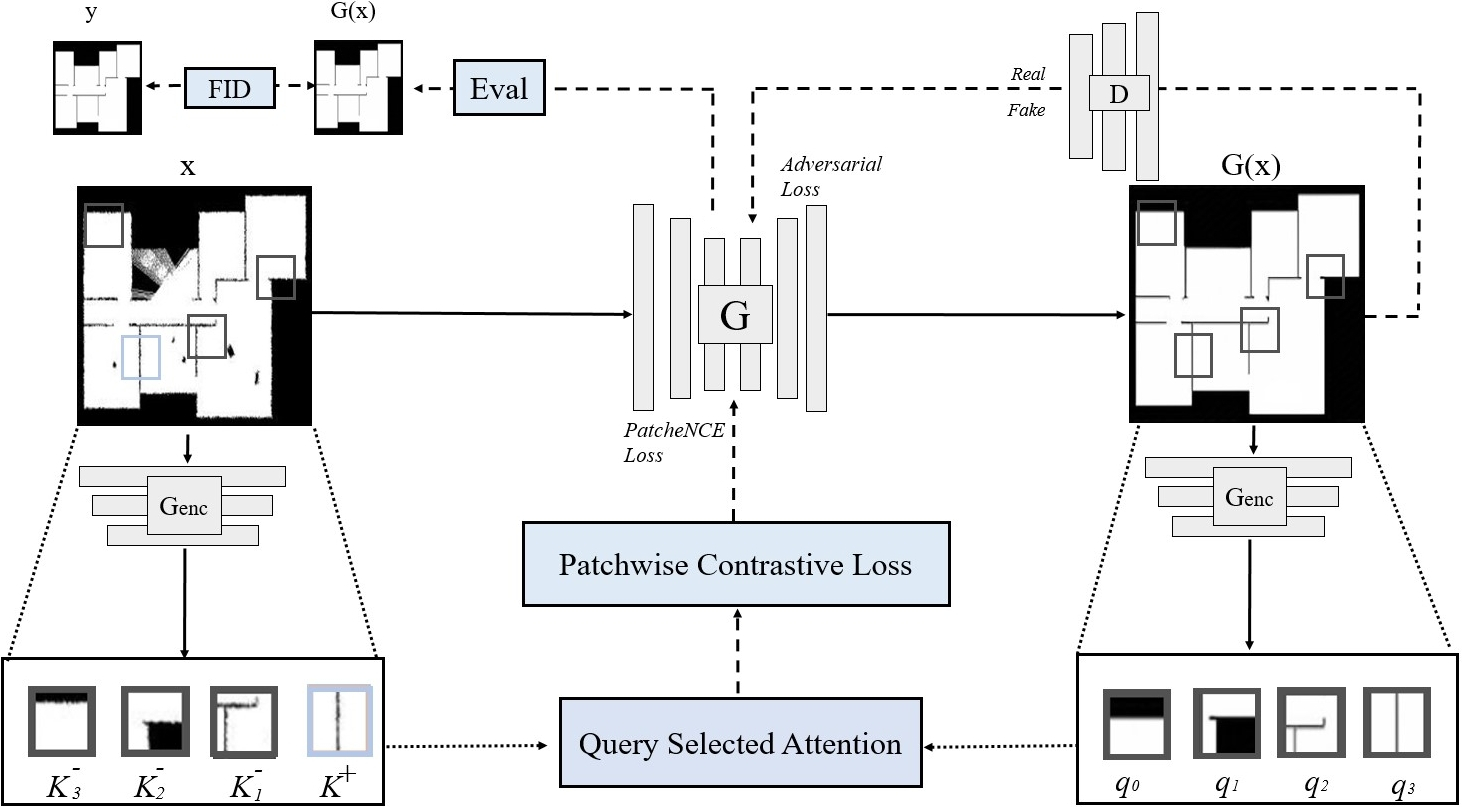}
    \caption{Training procedure for 2D occupancy grid cleaning through a GAN. $I_x$ is a training sample generated through our DRL data generator. $G(x)$ is a predicted clean variant of $x$. We use a Query-selected attention module \cite{qsa} for selection of $K^-$, $K^+$ and anchor points $q$ to compute PatchNCE loss from \cite{cut}, applied to Generator $G$. A PatchGAN discriminator\cite{patchgan} is used to judge the translation and produce an adversarial loss, also applied to $G$. After translation, images are evaluated for quality of translation through FID \cite{FID}.}
    \label{fig:model_diagram}
\end{figure*}

We treat the task of occupancy grid cleaning as an Image-to-Image, or OGM-to-OGM translation task (erroneous $\longrightarrow$ clean) and train a GAN model to learn this mapping. The objective of an I2I GAN is to translate images from input domain $X \subset \mathbb{R}^{H \times W \times 3}$
to resemble like images in output domain $Y \subset \mathbb{R}^{H \times W \times 3}$. In our approach, domain $X$ represents a database of erroneous OGMs, and domain $Y$ represents a database of pixel perfect OGMs. 
We represent an OGM as OGM[$H \times W$] akin to a matrix [$H \times W$] in structure. We use this  to justify our methodology of training an I2I GAN to operate on OGM data. 
The fundamental architecture of an I2I-GAN includes the usage of two neural networks: a Generator $G$ and a Discriminator $D$. GANs are successful in the task of Image-to-Image translation through an adversarial loss \cite{GANs}

\begin{equation}
\begin{aligned} 
\mathcal{L}_{GAN}(G, D, X, Y) = E_{y \sim Y} \log D(y) \\
+ E_{x \sim X} \log (1-D(G(x)))
\end{aligned} 
\end{equation}

\noindent The Adversarial loss encourages the generator to produce samples that are indistinguishable from samples within training the data.

The Generator $G$ in our GAN is based off ResNet \cite{resnet-gen} divided into $G_{enc}$ and $D_{enc}$. $G$ aims to take a sample of data from domain $X$ and modify it to more closely match the distribution of data in domain $Y$. We employ a PatchGAN \cite{patchgan} discriminator to judge this translation during the training process.

To learn this mapping function, we use a self-supervised patchwise contrastive loss (PatchNCE) \cite{cut}. 

\begin{equation}
\begin{aligned}
\mathcal{L}_{{con}} = -\log \left[ \frac{\exp(q \cdot k^+/\tau)}{\exp(q \cdot k^+/\tau) + \sum_{i=1}^{N-1} \exp(q \cdot k^-/\tau)} \right]
\end{aligned}
\end{equation}

\noindent $G_{enc}$ establishes similar features by comparing cross-domain patches between data samples in domains $X$ and $Y$. Here, $q$ represents an anchor feature, $k^+$ is a single positive, and $k^-$ are $(N - 1)$ negatives. $q$ locates features randomly in generated image $G(i_x)$, and the positive $k^+$ mirrors the location in input image $I_x$. $\tau$ denotes a temperature hyper-parameter. 

The Generator $G$ is trained to learn a unilateral mapping function from the gradient of $L_{\text{con}}$ applied on the anchor point $q$, separated from $k^+$ and $k^-$. We opt for a contrastive loss as apposed to a cycle-consistency based loss due to the constraints of cycle consistency loss on geometric translations which we require for linear/angular error correction.

PatchNCE initialises patches by randomly selecting features for $q$, $K^+$ and $K^-$. This random selection ignores any consideration of domain specific characteristics, potentially leading to inconsistent changes imposed by $L_{con}$ on generator $G$.

To address this issue, we incorporate a Query Selected Attention (QSA) module \cite{qsa} for selecting significant features. QSA aims to improve the selection of anchor point $q$, positive $k^+$ and negatives $k^-$ and applies $L_\text{con}$ across both domains to improve model efficiency. 

QSA reshapes a selected feature $F_x \in \mathbb{R}^{HW \times C}$ by its  transposed $K \in \mathbb{R}^{C \times H \times W}$ into a 2D matrix $Q \in \mathbb{R}^{HW \times C}$
and parses it through a softmax function to provide a global attention matrix $\text{Ag} \in \mathbb{R}^{HW \times HW}$. Significant features are identified by measuring the entropy $H_g$ in each  row of $A_g$ through the equation 

\begin{equation}
\begin{aligned}
H_g(i) = -\sum_{j=1}^{HW} \text{Ag}(i, j) \log \text{Ag}(i, j)
\end{aligned}
\end{equation}

\noindent Significant features are identified by finding the rows in $A_g$ with the highest entropy values $H_g$. The smallest $N$ rows are then used to identify the features in $I_x$ for the global attention matrix $AQS \in \mathbb{R}^{N \times HW}$.

The local attention matrix is used to capture interactions within local regions and reduce computational cost. It measures similarities between neighbouring queries and keys and is formulated similarly to global attention but is calculated on the queries themselves. Query matrix $Q_l \in \mathbb{R}^{HW \times C}$ is multiplied by the local key matrix $K_l \in \mathbb{R}^{HW \times w^2 \times C}$ and passed through a softmax function to provide the local attention matrix $A_l \in \mathbb{R}^{HW \times w^2}$. The entropy $H_l$ is computed for each row through the equation

\begin{equation}
\begin{aligned}
    H_l(i) = -\sum_{j=1}^{w^2} Al(i, j) \log Al(i, j)
\end{aligned}
\end{equation}

\noindent $A_{QS}$ is determined the same way for local attention as it was for global attention. $A_{QS}$ is used to determine significant features from $I_x$ for the PatchNCE loss to improve the translation efficiency and performance of the GAN.

\section{Model Learning and Experiment Results}






\subsubsection{Network Training}

Our OGM-to-OGM GAN model was trained on 75,000 samples of unpaired erroneous and pixel perfect OGMs on an Nvidia A100 for 100 epochs. Batch size was set to of 1 for stability. Learning rate was initially set to $ 2 \times 10^{-4}$ and lineally decayed to 0 over the last 50 epochs. The Adam optimiser was used with $\beta$ values of $\beta1 = 0.5$ and $\beta2 = 0.999$. The evaluation of the model's translation success was conducted using Fréchet Inception Distance \cite{FID}. We utilise the QSA module to locate features for the PatchNCE loss, applied to the last two layers of $G_{enc}$. The number of features selected for the PatchNCE loss was set to 256. The PatchGAN discriminator, evaluating the performance of the generator, consists of three layers of down-sampling and two convolutonal layers, and employs a Leaky ReLU activation function.

The complete objective of our OGM-to-OGM GAN model for occupancy grid cleaning is summarised as:

\begin{equation}
\begin{aligned}
L_G = L_{\text{adv}} + L_{\text{con}}^X + L_{\text{con}}^Y
\end{aligned}
\end{equation}

\noindent Where $L_\text{adv}$ represents the adversarial loss from the discriminator $D$, $L_{\text{con}}^X$ denotes the PatchNCE loss, and $L_{\text{con}}^Y$ signifies the identity loss. The identity loss uses $K^+$ and $K^-$ from image $I_y$ within domain $Y$, while $q$ represents anchor points from the generated image $G(I_y)$. The parallel contrastive loss and identity loss ensure similar features between $G(I_y)$ and $I_y$, inhibiting $G$ from altering $I_y$.

\subsection{Real-time SLAM: Haslegrave Dataset}

\newcolumntype{C}[1]{>{\centering\arraybackslash}m{#1}}

\begin{table*}[ht]
    \centering
    \caption{OGM Accuracy Comparison for Real-Time SLAM}
    \label{tab:iou_comparison}
    \begin{tabular}{lcccccc}
        \toprule
        \textbf{SLAM} & \textbf{Type} & \textbf{Unoccupied IoU $\uparrow$} & \textbf{Occupied IoU $\uparrow$} & \textbf{Baseline Map} & \textbf{GAN Map} \\
        \midrule \vspace{5pt}
        \multirow{2}{*}{\centering Gmapping} & Baseline & 0.2105 & 0.0281 & \multirow{2}{*}{\includegraphics[width=0.1\textwidth]{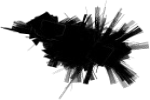}} & \multirow{2}{*}{\includegraphics[width=0.1\textwidth]{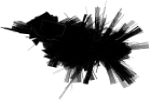}} \\
                                             & GAN      & 0.2092 & 0.0247 & & \vspace{5pt}\\
        \midrule \vspace{5pt}
        \multirow{2}{*}{\centering RTAB (ICP)} & Baseline & 0.4738 & 0.0268 & \multirow{2}{*}{\includegraphics[width=0.1\textwidth]{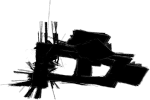}} & \multirow{2}{*}{\includegraphics[width=0.1\textwidth]{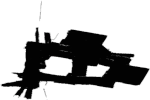}} \\
                                               & GAN      & 0.4974 & 0.0297 & & \vspace{5pt}\\
        \midrule \vspace{5pt}
        \multirow{2}{*}{\centering Hector} & Baseline & 0.2417 & 0.0293 & \multirow{2}{*}{\includegraphics[width=0.1\textwidth]{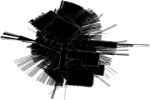}} & \multirow{2}{*}{\includegraphics[width=0.1\textwidth]{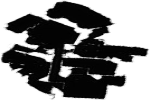}} \\
                                           & GAN      & 0.2804 & 0.0533 & & \vspace{5pt}\\
        \midrule \vspace{5pt}
        \multirow{2}{*}{\centering SLAM-Toolbox} & Baseline & 0.4525 & 0.0286 & \multirow{2}{*}{\includegraphics[width=0.1\textwidth]{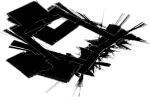}} & \multirow{2}{*}{\includegraphics[width=0.1\textwidth]{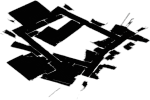}} \\
                                                 & GAN      & 0.4920 & 0.0346 & & \vspace{5pt}\\
        \midrule \vspace{5pt}
        \multirow{2}{*}{\centering F-LOAM} & Baseline & 0.5307 & 0.0382 & \multirow{2}{*}{\includegraphics[width=0.1\textwidth]{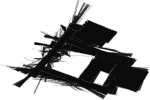}} & \multirow{2}{*}{\includegraphics[width=0.1\textwidth]{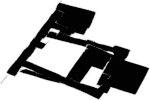}} \\
                                           & GAN      & 0.6134 & 0.0192 & & \vspace{5pt}\\
        \midrule \vspace{5pt}
        \multirow{2}{*}{\centering HDL} & Baseline & 0.5901 & 0.0262 & \multirow{2}{*}{\includegraphics[width=0.1\textwidth]{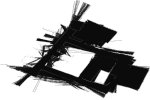}} & \multirow{2}{*}{\includegraphics[width=0.1\textwidth]{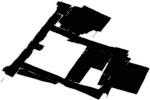}} \\
                                        & GAN      & 0.6888 & 0.0533 & & \vspace{5pt}\\
        \midrule \vspace{5pt}
        \multirow{2}{*}{\centering TT-OGM (Ours)} & Baseline & 0.6110 & 0.0281 & \multirow{2}{*}{\includegraphics[width=0.1\textwidth]{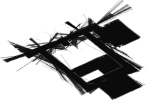}} & \multirow{2}{*}{\includegraphics[width=0.1\textwidth]{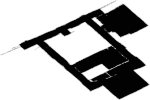}} \\
                                                  & GAN      & \textbf{0.8277} & \textbf{0.1149} & & \vspace{5pt}\\

        \bottomrule
    \end{tabular}
    
\end{table*}

We demonstrate our SLAM on real-world data of the Haslegrave building at Loughborough University. This dataset contains LiDAR recordings of a walk-through of the building. The building is large and several complex movements and turns were made during mapping in attempt to capture difficult to map data. LiDAR was captured from a singular Velodyne VLP-16 LiDAR scanner.

We evaluate our SLAM by comparing output map with a ground truth sample through IoU (intersect over union). The ground truth sample is a too-scale floor plan of the site created by an unaffiliated third party.

We demonstrate results of our SLAM as compared to other well known SLAM algorithms and provide aggregated testing of each SLAM with and without our GAN module.The 3D SLAM algorithms demonstrated are constructing 2D OGMs through our proposed Transformation and Translation process. 

The IOU calculation is done for both occupied and unoccupied space. Unexplored space is not taken into consideration as this differed too greatly between different samples. The ground truth image was cardinally aligned with the output of each SLAM, no other changes were made. Results are shown in Table \ref{tab:iou_comparison}. We also provide a higher resolution of the output of all SLAM algorithms as represented in ROS as occupancy grid maps. This is available in Table \ref{tab:real-time-results}

The estimated OGM created by our method and other 3D LiDAR-odometry algorithms using our pipeline outperform the traditional methods used for OGM creation. This demonstrates the viability and potential of our SLAM in large scale complex environments. To produce these results the exact same data sample was played, no changes were made to the 2D SLAM algorithms used for comparison. 

 TT-OGM with our GAN model produced the most accurate result as compared to the ground truth sample. The other SLAM methods, apart from Gmapping, also achieved a higher IoU with the GAN model correcting errors. The 2D SLAM algorithms (Gmapping, RTAB, Hector, SLAM-toolbox) were unsuccessful in producing a usable map of the scene. Despite achieving higher accuracy with the GAN the result is still unusable. The 3D SLAM methods (F-LOAM and HDL) were successful through our Transformation and Translation pipeline, results were further improved with our GAN error correction. 
The results for occupied accuracy are noticeably much lower than unoccupied, this is due to the constraint of comparing a SLAM map to a floor plan. Several features in the building are not labeled as occupied within floorplan whereas every occupied space is mapped in the SLAM. We decide to show these results as every SLAM scored similarly on this metric, our SLAM achieved the highest correct occupied IoU by more than double the second. We believe this proves the reliability of our method.

\subsection{Real-time Performance Evaluation}

\begin{table}[H]
    \centering
    \caption{Computational Time Comparison for OGM Creation on Embedded System.}
    \label{tab:time_comparison}
    \begin{tabular}{lcc}
        \toprule
        \textbf{SLAM} & \textbf{Type} & \textbf{Time (seconds) $\downarrow$} \\
        \midrule
        \multirow{2}{*}{Gmapping} & Baseline & 5.24 \\
                                  & GAN      & -- \\
        \midrule
        \multirow{2}{*}{RTAB (ICP)}     & Baseline & \textbf{1.01} \\
                                  & GAN      & 5.13 \\
        \midrule
        \multirow{2}{*}{Hector}   & Baseline & 2.00 \\
                                  & GAN      & -- \\
        \midrule
        \multirow{2}{*}{SLAM-Toolbox} & Baseline & 5.00 \\
                                      & GAN      & 13.15 \\
        \midrule
        \multirow{2}{*}{F-LOAM}   & Baseline & 1.17 \\
                                  & GAN      & 4.84 \\
        \midrule
        \multirow{2}{*}{HDL}      & Baseline & 1.41 \\
                                  & GAN      & 4.83 \\
        \midrule
        \multirow{2}{*}{TT-OGM (Ours)} & Baseline & 1.12 \\
                                       & GAN      & \textbf{4.81} \\
        \bottomrule
    \end{tabular}
\end{table}

A potential limitation of adding a GAN model to SLAM is the computational cost. In most circumstances these SLAM algorithms are running on embedded systems in which computational availability is limited. We demonstrate the computational time for a SLAM map to be constructed with and without our GAN model. We note that this is judging the time to construct an OGM, not localisation. These tests were produced on a Jetson Orin NX (16GB) embedded system. Results are available in Table \ref{tab:time_comparison}

\begin{table*}[ht]
\caption{Results of various SLAM algorithms on the Haslegrave Dataset.}
\centering
\begin{tabular}{|c|c|c|c|}
\hline
\textbf{Gmapping} & \textbf{RTAB (ICP)} & \textbf{Hector Mapping} & \textbf{SLAM-Toolbox} \\
\hline
\adjustbox{margin=3pt}{\includegraphics[width=0.2\textwidth, height=0.2\textwidth]{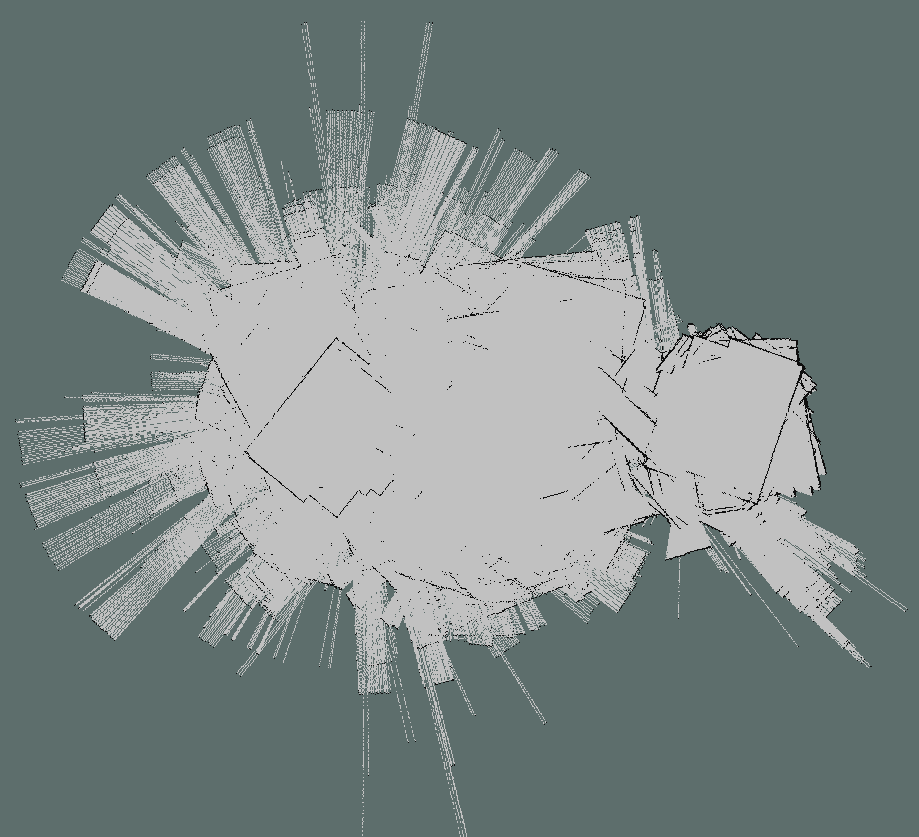}} & \adjustbox{margin=3pt}{\includegraphics[width=0.2\textwidth, height=0.2\textwidth]{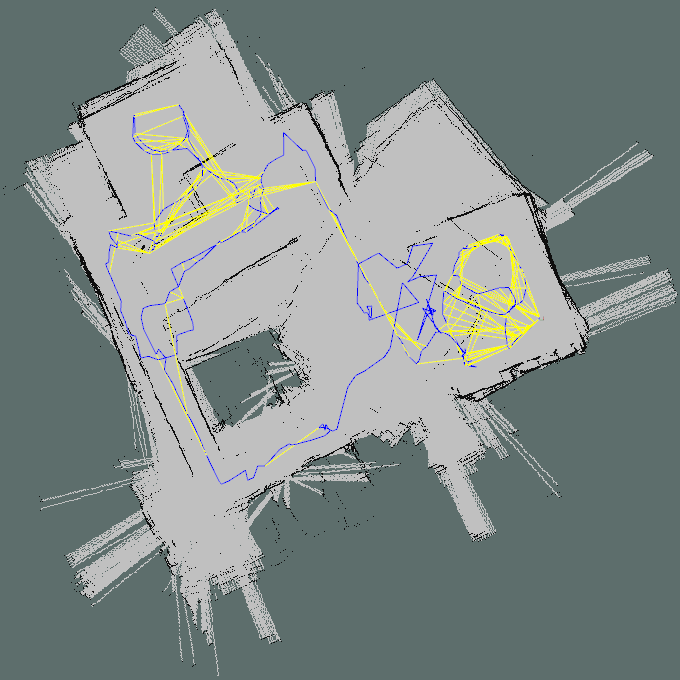}} & \adjustbox{margin=3pt}{\includegraphics[width=0.2\textwidth, height=0.2\textwidth]{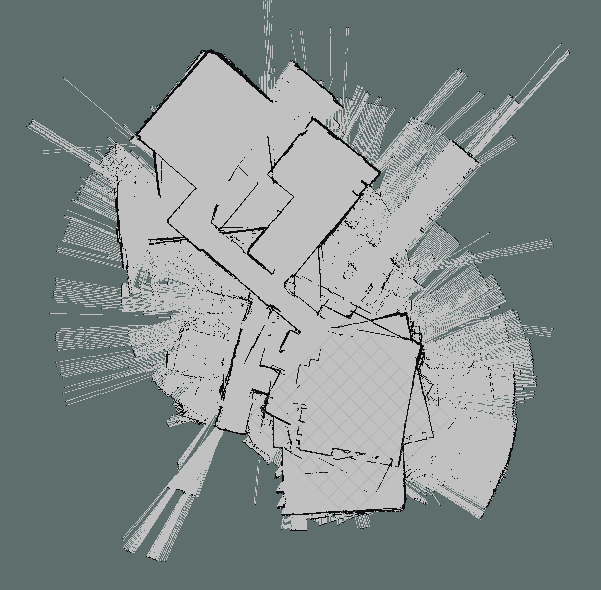}} & \adjustbox{margin=3pt}{\includegraphics[width=0.2\textwidth, height=0.2\textwidth]{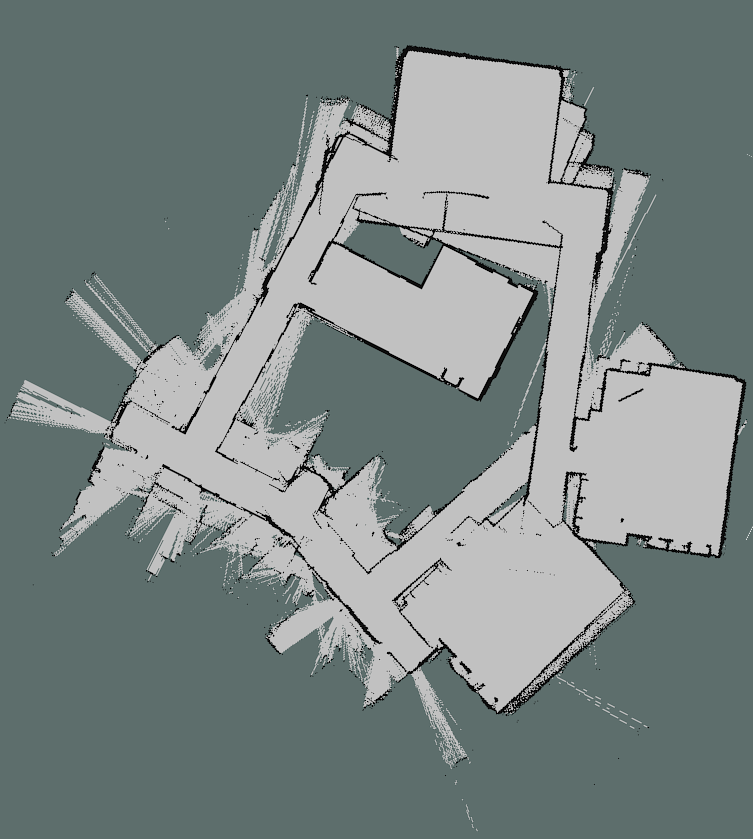}} \\
\hline
\textbf{F-LOAM} & \textbf{HDL} & \textbf{TT-OGM (No GAN)} & \textbf{TT-OGM (GAN)} \\
\hline
\adjustbox{margin=3pt}{\includegraphics[width=0.2\textwidth, height=0.2\textwidth]{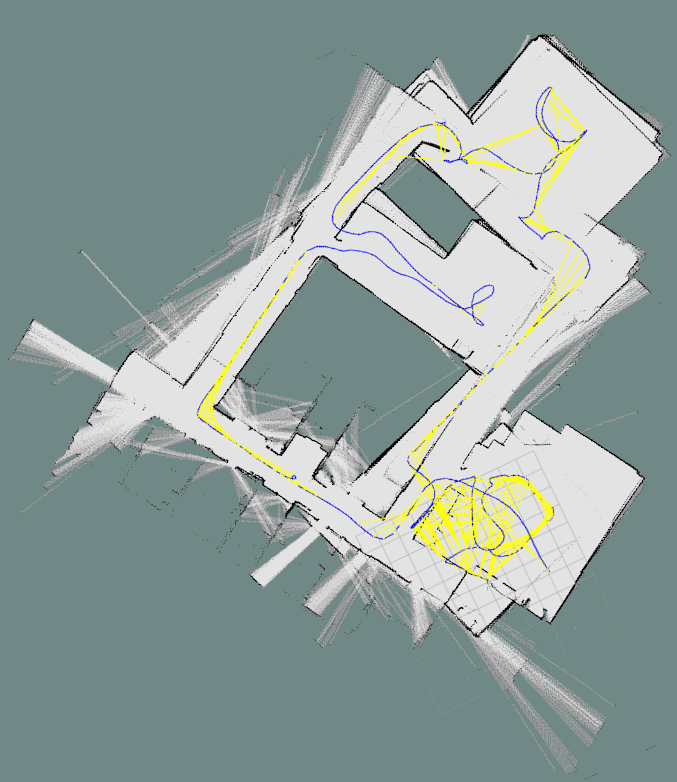}} & \adjustbox{margin=3pt}{\includegraphics[width=0.2\textwidth, height=0.2\textwidth]{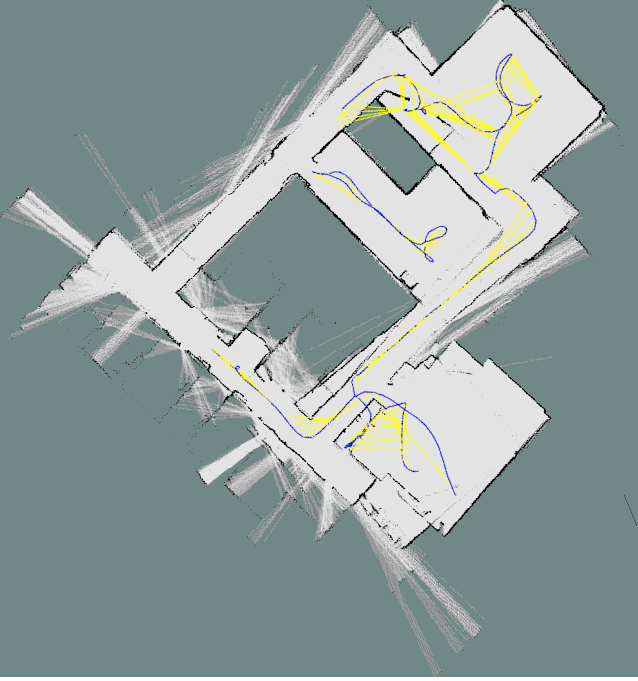}} & \adjustbox{margin=3pt}{\includegraphics[width=0.2\textwidth, height=0.2\textwidth]{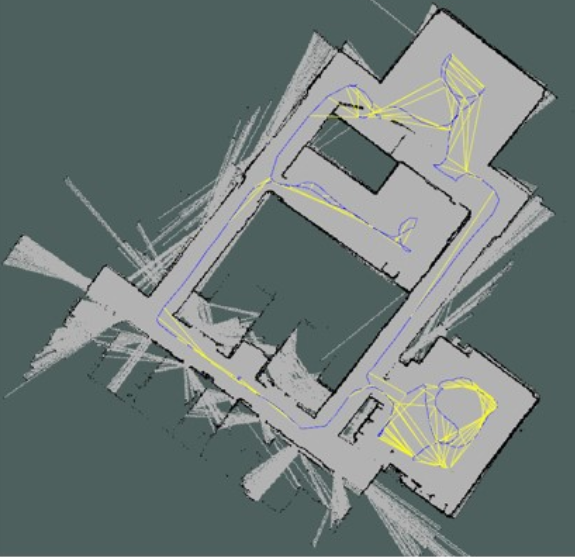}} & \adjustbox{margin=3pt}{\includegraphics[width=0.2\textwidth, height=0.2\textwidth]{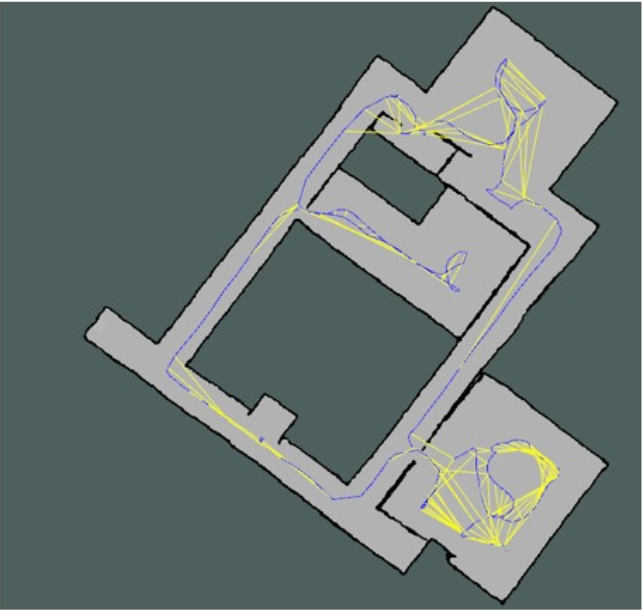}} \\
\hline
\end{tabular}
\label{tab:real-time-results}

\end{table*}
In these results, Gmapping and Hector SLAM were unable to use the GAN model in real-time. This was due to the SLAM using all available resources which could solved with better optimisation. To capture results for comparison for these two methods, the GAN model was ran on only the last frame of the map construction in the dataset. The viable SLAM methods for complex OGM creation (F-LOAM, HDL, and Ours) all saw increased time, this however still is reasonable for real-time performance. We note that these times are the average of the entire mapping session for a large environment. In smaller environments OGM creation with and without GAN are much faster.







\subsection{Observation Completion and Error Deletion}

\newcolumntype{C}{>{\centering\arraybackslash}m{0.15\textwidth}} 
\newcolumntype{L}{>{\centering\arraybackslash}m{0.07\textwidth}}
\begin{table*}[ht]
    \centering
     \captionsetup{justification=centering} 
    \caption{Predictions made on samples from the Radish dataset \cite{radish}: (a) SDR Site B by Andrew Howards, (b) Fort AP Hill by Andrew Howards, (c) Logwood by Nick Roy, (d) Intel Lab by Dieter Fox, (e) Albert-b Laser Vision (Freiburg).}
     \label{tab:predictions_generator_radish}
    \begin{tabular}{|L|*{5}{C|}}
        \hline
        & (a) & (b) & (c) & (d) & (e) \\
        \hline
        Baseline & 
        \includegraphics[width=0.15\textwidth]{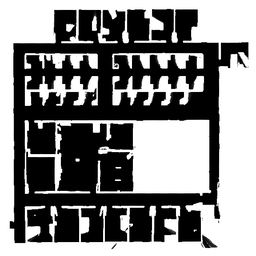} &
        \includegraphics[width=0.15\textwidth]{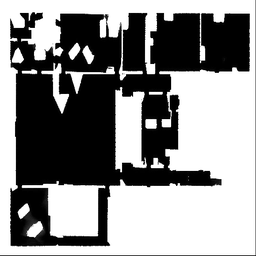} &
        \includegraphics[width=0.15\textwidth]{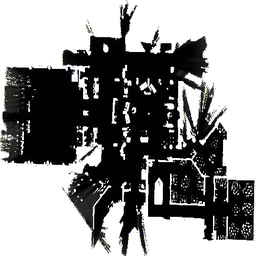} &
        \includegraphics[width=0.15\textwidth]{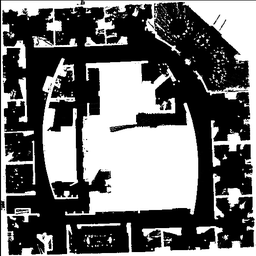} &
        \includegraphics[width=0.15\textwidth]{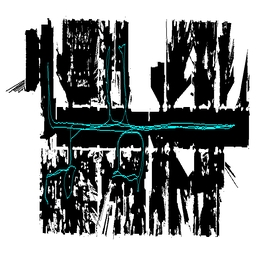} \\
        \hline
        
        Ours & 
        \includegraphics[width=0.15\textwidth]{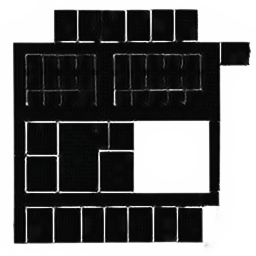} &
        \includegraphics[width=0.15\textwidth]{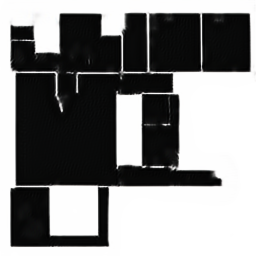} &
        \includegraphics[width=0.15\textwidth]{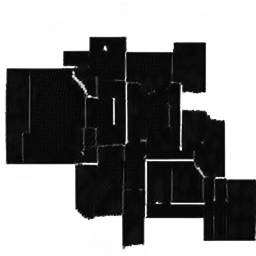} &
        \includegraphics[width=0.15\textwidth]{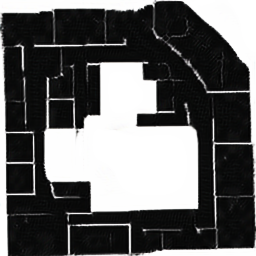} &
        \includegraphics[width=0.15\textwidth]{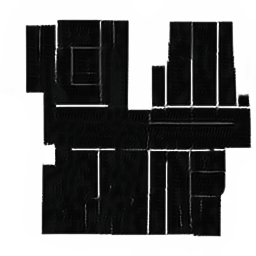} \\
        \hline
        Ours in ROS & 
        \includegraphics[width=0.15\textwidth]{radish/OccMap/rad1.png} &
        \includegraphics[width=0.15\textwidth]{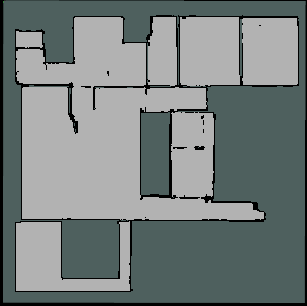} &
        \includegraphics[width=0.15\textwidth]{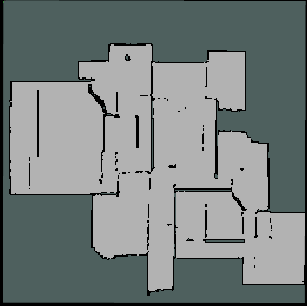} &
        \includegraphics[width=0.15\textwidth]{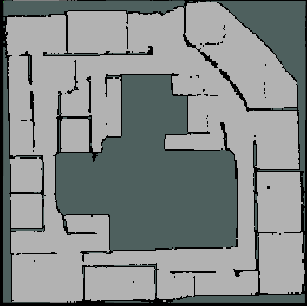} &
        \includegraphics[width=0.15\textwidth]{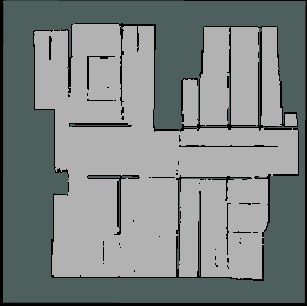} \\
        \hline
    \end{tabular}

        \vspace{10pt} 
    
    \label{tab:radish_predictions}
\end{table*}

The Radish dataset is a well known 2D SLAM dataset. The typical criteria for assessing a SLAM through the radish dataset is through comparing results by producing a map from the LiDAR data within each sample. We demonstrate these results to show the success of our method for the task of observation completion and error deletion. With these, we aim to prove our method on a variety diverse, large, and famous indoor SLAM scenes. 

Our criteria of observation completion and error deletion is a complex task to evaluate on this dataset. As our method is new and unique there exists no ground truth error free, complete samples. However there is noticeable visual improvement in mapping quality and completeness. We invite readers to make their own judgements on our results compared to baseline. 

We build our results by attaching our GAN model as a final stage to the input SLAM of Gmapping. Results are available in Table \ref{tab:predictions_generator_radish}.

\begin{figure}[H]
    \centering
    \includegraphics[width=0.2\textwidth]{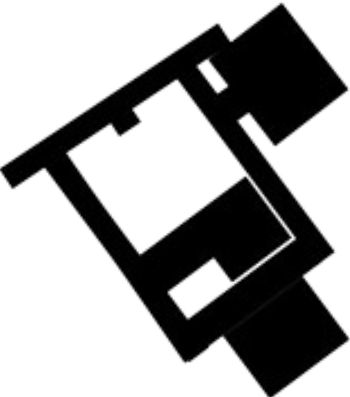}
    \caption{Ground Truth Image of the Haslegrave SLAM Map used for IoU comparison}
    \label{fig:ground_truth}
\end{figure}








\section{Conclusions}

In this paper, we introduced a new and novel approach to OGM with our Transformation \& Translation Occupancy Grid Mapping. Our method addresses the significant limitations of traditional OGM in large, complex environments with dynamic movement. By adapting state of the art 3D-LiDAR pose estimation techniques typically used in 3D SLAM and utilising GANs for observation completion and error correction we have significantly enhanced the quality and usability of 2D OGMs. 

Our Transformation \& Translation pipeline enables 3D SLAM algorithms to directly produce 2D OGMs, which far surpass 2D SLAM in terms of accuracy and quality. This approach allows for accurate 2D mapping in complex scenes, effectively bridging the gap between 2D and 3D SLAM. 

Our Deep-learning based SLAM error correction leverages GANs to detect and correct errors and incomplete observations in real-time. This not only improves mapping accuracy but also ensures that produced maps are visually cleaner. This benefits other domains that could benefit from quickly produced distance-accurate floor plan diagrams to consider the usage of our SLAM.

To address the scarcity of large, diverse datasets for training deep learning models for SLAM improvement, we introduced a novel data generation process using DRL. Our method enables the creation of realistic and varied 2D SLAM errors at scale, providing the means to generate an almost unlimited amount of data. We hope our data can be utilised for further advancements in deep learning methodologies for OGM. 

We prove our TT-OGM framework using real-world data collected at Loughborough University, and on famous examples of OGMs from the Radish dataset. These tests demonstrate our method's effectiveness in complex and dynamic environments. Our results demonstrate superiority over other existing OGM algorithms.


       




\end{document}